%% file: root.tex
\documentclass[letterpaper, 10 pt, conference]{ieeeconf}  

\IEEEoverridecommandlockouts                              

\usepackage{graphics} 
\usepackage{epsfig} 
\usepackage{times} 
\usepackage{amsmath} 
\usepackage{amssymb}  
\usepackage{xcolor}
\usepackage{hyperref}

\usepackage{threeparttable} 
\usepackage{tabularx}
\usepackage{multirow} 
\usepackage{booktabs} 
\usepackage[autostyle]{csquotes}
\usepackage{fancyhdr}
\fancypagestyle{firstpage}{ 
    \fancyhf{} 
    \fancyhead[L]{Accepted to IEEE International Conference on Robotics and Automation (ICRA) 2025} 
}

\title{\LARGE \bf
Hybrid Deep Reinforcement Learning for Radio Tracer Localisation in Robotic-assisted Radioguided Surgery
}

\author{
        Hanyi Zhang$^{1}$,
        Kaizhong Deng$^{1}$,
        Zhaoyang Jacopo Hu$^{2}$,
        Baoru Huang$^{1}$,
        Daniel S. Elson$^{1}$ %
\thanks{$^{1}$Hamlyn Centre for Robotic Surgery, Department of Surgery and Cancer, Imperial College London.}%
\thanks{$^{2}$Department of Mechanical Engineering, Imperial College London.}%
\thanks{ Corresponding email: {\tt\small daniel.elson@imperial.ac.uk}}%
}

\begin{document}

\maketitle
\thispagestyle{firstpage}
\pagestyle{empty}

\begin{abstract}

Radioguided surgery, such as sentinel lymph node biopsy, relies on the precise localization of radioactive targets by non-imaging gamma/beta detectors. Manual radioactive target detection based on visual display or audible indication of gamma level is highly dependent on the ability of the surgeon to track and interpret the spatial information. This paper presents a learning-based method to realize the autonomous radiotracer detection in robot-assisted surgeries by navigating the probe to the radioactive target. We proposed novel hybrid approach that combines deep reinforcement learning (DRL) with adaptive robotic scanning. The adaptive grid-based scanning could provide initial direction estimation while the DRL-based agent could efficiently navigate to the target utilising historical data. Simulation experiments demonstrate a $95\%$ success rate, and improved efficiency and robustness compared to conventional techniques. Real-world evaluation on the da Vinci Research Kit (dVRK) further confirms the feasibility of the approach, achieving an $80\%$ success rate in radiotracer detection. This method has the potential to enhance consistency, reduce operator dependency, and improve procedural accuracy in radioguided surgeries.

\end{abstract}

\section{INTRODUCTION}
Prostate cancer is one of the most common cancers among men in the UK, with more than 52,000 men diagnosed with prostate cancer every year~\cite{gnanapragasam2022urinary}. The histological status of pelvic lymph nodes (LN) is considered a significant prognostic factor in prostate cancer~\cite{gervasi1989prognostic}. According to the European Association of Urology guidelines, extended pelvic lymph node dissection (ePLND) remains a preferred tool for staging~\cite{mottet2021eau}. However, the ePLND increases the chance of postoperative complications and is complex and time-consuming~\cite{fossati2017benefits}. Sentinel node biopsy represents a promising methodology with potentially lower unnecessary morbidity and more accurate staging by removing the cancerous LNs outside the standard lymphadenectomy regime~\cite{mehralivand2018sentinel}. To aid sentinel node biopsy within the complex abdominal anatomy, patients may be injected with a cancer-targeting drug before surgery and then the LNs are identified by searching for the radiotracer signal (\enquote{hot spot}) with a nonimaging gamma probe. 

\input{image/problemstatement}

To minimize the invasiveness of surgery, there has been a transition from open procedures to laparoscopic techniques~\cite{van2013sentinel}, and more recently, to robot-assisted laparoscopic procedures~\cite{van2016revolutionizing,kleinjan2014optimisation}. Although various intraoperative radioguidance modalities have been explored, gamma probes with acoustic and numerical feedback remain the primary choice for intraoperative guidance~\cite{meershoek2019robot}. To enhance the accuracy of cancer and sentinel LN detection in robot-assisted procedures, drop-in gamma probe technology has been developed, offering improved maneuverability and precision~\cite{van2016revolutionizing, junquera2023drop}. However, relying on the gamma-level display or audible feedback is not an intuitive way to identify sentinel nodes and places a high demand on the surgeon's experience with the risk of overlooking latent hot spots.

Autonomy in surgery has been proven as a promising way to reduce the dependence on surgeon experience, enhance efficiency, and improve consistency~\cite{hu2023towards,chen2024human,zheng2024user,weiser2008estimation, attanasio2021autonomy, han2022systematic, saeidi2022autonomous, AutoSurgLap}. Over the past few years, autonomous systems were successfully applied to various surgical tasks such as suturing~\cite{pedram2017autonomous}, tissue manipulation~\cite{shin2019autonomous}, laparoscopy control~\cite{AutoSurgLap}, and tumour ablation~\cite{hu2018semi}. In addition, an imitation learning-based approach has been proposed to detect the radioactive target~\cite{deng2023deep}. 
However, this approach faces challenges in efficiency and responsiveness to real-time signal feedback.

Achieving full autonomy in complex and dynamic environments remains challenging due to the need for real-time decision-making. Deep Reinforcement learning (DRL) has emerged as a powerful tool, enabling autonomous robotic systems to learn and adapt through trial and error interactions within the environment~\cite{singh2022reinforcement,zhang2021reinforcement}. This allows iterative performance improvement without the need for human interaction~\cite{varier2020collaborative,xu2021surrol,datta2021reinforcement}. This approach is particularly well-suited to our task of radio tracer detection, where the robot has to adjust its actions to reach a target with an unknown position based on noisy signals from the environment.  

This work presents a hybrid approach based on adaptive scanning and deep reinforcement learning. The adaptive scanning initialises the starting condition by pointing the gamma probe towards the hot spot, achieved via systematic scanning while DRL agent effectively manipulates the probe to approach to the hot spot, thus localizing the radioactive spot. The hybrid approaches are proved to be an accurate, efficient and robust method for radiotracer detection compared with the separate components of the method. 
This was further validated through real-world experiments on the da Vinci Research Kit (dVRK)~\cite{kazanzides2014open, d2021accelerating}, which achieved a high success rate in hot spot detection.
This study provides an automated radiotracer detection approach, as well as laying the foundation for unidimensional signal localization tasks. 
Critically, the methodology presents the potential to mitigate operator-dependent variability, subsequently improving both the accuracy and operational efficiency of robot-assisted surgical interventions.

\section{PROBLEM STATEMENT}

\textbf{Task Description.} In radioguided surgery, the surgeon should manoeuvre the probe to localize the hot spot as shown in \autoref{fig:problemstatement}. The gamma probe responds to the radioactivity within a defined conical sensitivity volume in front of the detector and provides a real-time intensity measurement in counts per second (CPS). The operator can adjust the direction of the gamma probe based on changes in the received signal to attempt to navigate in the direction of higher response. Once the locally highest CPS has been found, the hot spot can thus be localized. 

The radiotracer detection problem can be formulated as a target-reaching problem based on one-dimensional signal feedback. The probe is moved in the direction where the CPS signal increases until it reaches the vicinity of the LN where the highest CPS can be achieved, indicating a successful detection. The problem of robot-assisted radiotracer detection can therefore be modelled as a Markov Decision Process defined by the tuple \( (\mathbf{S}, \mathbf{A}, \mathbf{P}, \mathbf{R}) \), representing the state, action, process model, and corresponding reward.

\textbf{State Space.} The state space represents the surgeons' observations during the procedure, based on which the surgeon can decide the next action. This task requires more information to assist in decision-making compared with general target-reaching tasks where the position of the target is known. To mimic the natural detection process where surgeons use the temporal variation in CPS to estimate the target direction, current and previous probe signal information is included in the observation space. Therefore, the state space consists of two parts: (1) the current state of the robot and (2) the CPS and position of the gamma probe, including their current values as well as historical values from the previous several steps. It is denoted as $s_t = \{ \mathbf{S}_{robot}(t), \mathbf{S}_{probe(t-k:t)} (t)\}$, where \( \mathbf{S}_{robot} \) is the state of the robot such as joint positions and velocity, and \( \mathbf{S}_{probe(t-k:t)} \) is the information of the gamma probe including the signal and position, along with previous $k$ steps.

\textbf{Action Space.} The action space represents the surgeon's actions, which are translated into robot control commands for each joint. These actions control the robot’s joints to manipulate the probe. The action space \( \mathbf{A} \) can be defined as the set of control actions applied to the robot’s joints: $a_t = \{ \Delta \theta_1(t), \Delta \theta_2(t), \dots, \Delta \theta_6(t) \}$ where \( \Delta \theta_i(t) \) represents the incremental control applied to joint \( i \) at time \( t \).
The agent can adjust the robot pose based on the current state.

\textbf{Reward.} The goal of the procedure is to successfully detect the hot spot, meaning the gamma probe must reach the vicinity of the radiotracer source. 
Besides, this procedure should also ensure the safe movement of the probe.
Thus, it is designed to provide a dense reward based on the distance between the probe tip and the target for efficient probe targeting. To ensure safety, a penalty was added for moving too quickly, which helps prevent sudden or jerky movements. An additional bonus is given when the probe reaches a certain threshold distance from the target.

\input{image/framework}
\section{METHODOLOGY}

\subsection{DRL-Based Hybrid Approach}
This paper presents a hybrid methodology integrating DRL with adaptive scanning techniques to address critical challenges in hot spot localization.

While traditional adaptive scanning methods provide comprehensive grid-based exploration for highly reliable results, they often suffer from significant time constraints in the dynamic surgical environment, which makes it infeasible to be implemented in real surgical scenarios.
Conversely, pure DRL approaches encounter significant initial exploration constraints, making it challenging to estimate the target's approximate location during the early exploration phase, particularly when targets are positioned distant from the starting detection point.

Our proposed hybrid approach addresses these challenges by utilizing an adaptive scanning approach in the initial phase to provide directional guidance and accumulate early experience for DRL-based detection. 
This transforms the DRL optimization problem from broad directional exploration to refined directional adjustment.
With a limited amount of preliminary scanning, this method reduces the burden of DRL exploration, thus enhancing robustness while maintaining a balance between detection efficiency, robustness, and accuracy.

\subsection{Phase I: Adaptive Scanning}
In adaptive scanning Phase I, the gamma probe approached the target while continuously scanning the environment as shown in \autoref{fig:framework}. The adaptive scanning provided initial data for DRL as prior experience and adjusted the probe to a general target direction, thereby confining the target within a detectable range and enhancing the reliability of the method. At each step of scanning, the probe performed scans by varying its pitch and yaw angles within a range of $[-\alpha,\alpha]$, creating a two-dimensional $5\times5$ grid representing the distribution of radiation intensity. This grid was pre-processed with a Gaussian blur to reduce noise and help to identify the peak signal. The gamma probe was then reoriented towards this peak and the probe was moved in that direction. 
 
To improve efficiency and accuracy, the scanning range was progressively narrowed by a factor $\gamma$ after each step, allowing the probe to hone in more quickly on the exact LN position as $\alpha_{i+1}=\gamma\alpha_{i}$.
Based on previous experimental results, the initial scanning range was set as $\alpha_0 = 30^\circ$ and the optimal parameters were determined to be $\gamma = 0.8$.

In phase I, the target was typically confined to a narrower range (with $\alpha = 19^\circ$) after two rounds of scanning. Therefore, we set the base number of scans to two. If the peak value still appears at the edge of the grid after two rounds, additional scans are performed until the global peak direction is identified. The process stops once the peak no longer lies on the boundary, ensuring the target direction is accurately determined before transitioning to the next phase.
The adaptive scanning provides initial data for DRL as prior experience and adjusts the probe to a general target direction target, thereby confining the target within a detectable range and enhancing the reliability of the method.

\subsection{Phase II: DRL-Based Approach}
In phase II, we employed the Surgical Gym platform~\cite{schmidgall2024surgical}, an open-source framework for surgical robot learning, to train a DRL agent using Proximal Policy Optimization.

Since the location of the LN was initially unknown to the robot, the agent should rely on the current and historical data to locate it.
However, directly incorporating historical information into the state space may confuse the agent in understanding the relationship between inputs from different time steps. To address this problem, we developed an Angle Prediction Module to utilize historical data with temporal information to estimate a rough position of the target. A key component of this model is the Corrected Solid Angle Model, an analytical framework that mathematically describes the radioactivity response on the gamma probe. This model estimates the detected radiation fraction by considering both geometric attenuation and detection probability. By explicitly modelling the relationship between the detection angle and CPS with this model, the Angle Prediction Module enables estimation of the target's approximate position. The CPS was calculated by~\cite{hartl2015detection}:

\begin{equation}
    a \approx C_1 \cdot \frac{1}{2} \cdot \left(1 - \frac{1}{\sqrt{\frac{r^2}{d^2} + 1}}\right) 
    \cdot f(\alpha) + C_2,
\label{modelequation}
\end{equation}
and the scale function can be expressed as:
\begin{equation}
    f(\alpha) =
    \begin{cases} 
        1, & \text{if } \alpha \leq \arctan\left(\frac{2r}{l}\right) \\[8pt]
        \frac{2r }{l \tan(\alpha)}, & \text{if } \alpha > \arctan\left(\frac{2r}{l}\right) 
    \end{cases}.
\label{function_f}
\end{equation}
In these equations, $a$ represents the detected signal, $r$ is the effective radius of the gamma probe’s detection area, $d$ is the distance between the probe and the source, $l$ is the interactive length of the detector, and $\alpha$ is the detection angle, defined as the inclination angle between the probe and the source.

To align the radioactivity simulation more closely with the actual gamma probe response, we employed coefficients \( C_1 \) and \( C_2 \) that calibrate the idealized model to empirical measurements, effectively bridging the gap between theoretical predictions and real-world radiation detection.

Based on \autoref{modelequation}, the module used these data to fit the model parameters and estimate the angle of the target based on the initial guess of its position from adaptive scanning. The estimated relative angle was then fed into the state space. This approach enabled the robot to observe both the current CPS intensity and the detection angle, allowing it to navigate the probe in the direction where the signal strength increased, guiding it more accurately toward the target.

The reward function $\mathbf{R}$ was designed to guide the gamma probe $\mathbf{P}$ towards the radioactive source $\mathbf{T}$ with high precision and efficiency. The reward function combined dense rewards, potential rewards, speed penalty, and terminal rewards. The dense rewards were distance-based and intrinsic which can be expressed as: 
\begin{equation}
    R(s_t)= -\left|\mathbf{P} - \mathbf{T}\right|.
\end{equation} 

As the probe moves towards the target, the dense reward becomes too small to provide useful guidance. Therefore, the additional signal-based potential reward was introduced to compensate the incentive for the probe during the final stages of detection, which has been suggested to accelerate learning and lead to an optimal behaviour~\cite{devlin2012dynamic}. The potential rewards were based on the feedback signal as:
\begin{equation}
F(s_t, s_{t-1})=CPS(t) - CPS(t-1).
\end{equation} 

The terminal rewards were given when the target was successfully detected, defined as:
\begin{equation}
R_s(s_H) = 
\begin{cases} 
    1, &  |\mathbf{P}_H - \mathbf{T}| < 5\ mm \\
    0, & \text{otherwise}
\end{cases}
\end{equation}
The speed penalty was defined as:
\begin{equation}
    R_p(s_t) = - \lambda_q \left|\mathbf{\dot{q}}\right|, 
\end{equation}
which could punish joint velocities to encourage a smooth and steady approach to the target. The final reward function integrated these four rewards and penalties over the horizon $H$ which can be expressed as:
\begin{equation}
    G = \sum_{t=0}^{H-1} \left( R(s_t) + \lambda_f F(s_t, s_{t+1}) + \lambda_q R_p(s_t) \right) + \lambda_s R_s(s_H),
\end{equation}
where $\lambda_f$, $\lambda_q$, and $\lambda_s$ are hyperparameters.

\section{Experiment and Results}
Our experiments were conducted on both the simulation platform and the real-world system. We built a simulation environment with a surgical robot manipulator and a simulated radioactive source. We trained a deep reinforcement learning agent in this simulation by tuning reward function for optimized results. The trained agent was then evaluated in both simulation setting and real-world setting with dVRK.

\subsection{Simulation Environment Setup}
\input{image/simulationenv}

\subsubsection{\textbf{Robot Platform}}

In the simulation experiments, the da Vinci Patient Side Manipulator was trained and tested using Surgical Gym~\cite{schmidgall2024surgical}, a high-performance, GPU-based platform designed for surgical robot learning. Built on Isaac Gym~\cite{makoviychuk2021isaac}, Surgical Gym leverages GPU-accelerated physics simulations and reinforcement learning to provide significantly faster data collection rates compared to traditional CPU-based platforms. This speed improvement allows the simultaneous operation of thousands of robotic environments, making it possible to rapidly train models through Proximal Policy Optimization (PPO).

On this platform, the Patient Side Manipulator was controlled by a Proportional-Differential controller driving by a target joint-position which was the action space for the agent. The state of the robot included the joint positions and velocities. Tool tip positions could also be provided for the observation space. The platform allowed efficient testing of our hybrid algorithm in over 2,000 environments running in parallel, significantly reducing training time while ensuring robust model performance.

\subsubsection{\textbf{Radioactivity Modeling}}

To enhance realism, the gamma probe response to a real radioactive source was measured and modeled. The data were collected at varying distances and angles from the radiation source in a radiation-safe laboratory setting. A movable platform was used to support a real gamma radiation source, allowing movement only along horizontal and vertical axes. The gamma probe was positioned above the platform and initially centered in contact with a disc-shaped sealed radioactive Cobalt-57 gamma source. The platform moved vertically within a range of $[5, 45]\, mm$, collecting horizontal data at $5\,mm$ intervals. At each height, the platform moved horizontally within the range of $[0, 15]\, mm$, collecting data at $5\, mm$ intervals. Additionally, to capture data near the signal source where signal variations are larger, the platform moved only vertically within a range of $[1, 5]\, mm$, collecting horizontal data at $1\, mm$ intervals. For each position, 10 signal values were collected to obtain a mean value.

40 sets of data were collected, of which 36 were valid, while four sets were discarded because the signal was nearly unreadable due to excessive distance or extreme angles. The goal was to determine the optimal values of the model parameters \( l \), \( C_1 \), and \( C_2 \) in \autoref{modelequation}, which minimized the residual error between the predicted CPS values from the model and the actual observed data.
\input{image/modelfitting_fig}

The fitted parameters were as follows: the detector length $l$ was 26.50 mm, the scaling factor for the geometric term $c_1$ was 173.78, and the background offset $C_2$ was 0.29. The root mean square error (RMSE) of the fitted model is $2.78$. 
The $R^2$ value for the fitted model was $0.97$, indicating a high degree of correlation between the model predictions and the actual CPS data. The model fit to the data is shown in \autoref{fig:modelfit}.

\subsection{Reward Design}
The learning efficiency and performance of the radiotracer detection task were evaluated under different dense reward settings: a distance-based reward, a signal-based potential reward, and a reward with a combination of these. Two metrics were used to evaluate the training performance: the success rate and episode length, as shown in \autoref{fig:success-rate}. Both of the figures show that the composite reward achieves higher learning efficiency and stability compared to two individual reward functions. 

The distance-based reward exhibits a low success rate and slightly longer episode length. This is due to the lack of incentive to accurately locate the target in the final stage. Besides, the distance-based reward requires a longer time to complete the detection. 

The signal-based reward could effectively encourage the probe to locate the target in its vicinity precisely. However, the results show that it has a stronger fluctuation on its success rate as well as episode completion time during training, indicating its low training stability. This instability arises because the signal does not vary significantly in the initial stages, which leads to weak rewards and increases the risk of detection failure at the beginning. Moreover, this reward is dependent on the specific signal strength of the target, making it less compatible with varied signal levels as in the real clinical scenario. 

Therefore, balancing these two reward functions to provide stable incentives throughout the entire detection process is crucial for the training. Through experimentation, it was observed that setting the $\lambda_f=0.1$ could lead to a stable reward throughout the process, allowing for faster, smoother, and more efficient training.

\input{image/reward}
\input{tables/robustness}
\subsection{Performance Evaluation in Simulation}
To evaluate and compare the performance of the individual approaches and the hybrid approach, each approach was tested across 3,200 independent simulation runs (50 sets of tests with 64 environments running simultaneously) to ensure statistical significance. The performance of each approach is evaluated by the following metrics: 

\begin{itemize}
 \item \textbf{Success Rate}: Defined as probe localization within a $5\,mm$ distance threshold and maximum of 150 steps.
 \item \textbf{Average Steps}: Calculated as the mean number of steps to successfully locate the LN, excluding trials exceeding the 150-step limit.
 \item \textbf{Robustness}:  Evaluated by randomizing LN positions using normal distribution offsets of $30\,\text{mm}$, $50\,\text{mm}$, and $70\,\text{mm}$ standard deviations to assess performance consistency across varied target locations.
 \end{itemize}

\input{tables/performance}
The simulation result is shown in \autoref{tab:comparison}. The \textbf{Adaptive Scanning Approach} demonstrated a high success rate ($96\%$), indicating its effectiveness in locating LNs across various positions within the workspace. However, excessive scanning of each step makes it the least time-efficient. The \textbf{DRL Approach} significantly enhanced the efficiency with only $109$ steps taken to reach the target. However, it exhibited a low success rate and low robustness. For instance, when the randomization range of the target was large, there was a high chance that the target could not be detected initially, which led to the failure of the whole process.
This is caused by the poor initialisation of its orientation, which may be oriented pointing in the opposite direction to the target. The \textbf{DRL-based Hybrid Approach} combined the strengths of two methods and achieved an optimal balance among these three performance metrics. 
With the guidance of scanning results, the gamma probe could reach targets within a large range of positions through DRL training with high efficiency.

\subsection{Real-world Validation}

We evaluated the capability of the developed approaches in detecting the radioactive source in a real-world setting on the dVRK. The experimental setup is shown in \autoref{fig:experimentsetup}. A dVRK PSM system equipped with a ProGrasp EndoWrist instrument to grasp the gamma probe. A workstation was used to control the dVRK system, while another workstation was used to process the data and run the control method.

The target was placed in various positions within the workspace. The detection distance ranged from $30\,\text{mm}$ to $60\,\text{mm}$ and the angle ranged from $0$ to $60^{\circ}$. Due to the unavailability of radioactive sources within the dVRK operation area, a physical target with simulated radioactivity was employed in this experiment. The target's coordinates were first identified by using robot forward kinematics and then sent to the computer system to calculate the CPS. This consisted of a radioactive source-gamma probe interaction model including a simulated background noise. 

Only the hybrid approach was tested in real-world experiments because the other two individual approaches could not be feasibly implemented. For instance, the low efficiency takes an extremely long time to complete the scanning, while the DRL-based approach shows a significantly lower success rate than the hybrid approach. Therefore, the hybrid approach was recorded a total of ten times.

Experimental validation demonstrated the effectiveness of the hybrid approach, achieving an $80\%$ success rate in locating LNs.
Overall, it took an average of 96 steps to complete the detection procedure, which was $9.6\,\text{s}$. The robot could navigate to most of the targets despite the significantly varying locations, showcasing its robustness.

During the experimental trials, two failure cases were observed. In the first case, the probe approximately reached the target location but with a minor lateral deviation. The second trial demonstrated a more substantial navigational error, with the probe trajectory diverging significantly from the expected path.
These failures occurred when the target was placed farther away from the probe. Further investigation showed that the adaptive scanning failed to locate the target at the beginning, indicating that a robust adaptive scanning mechanism is critical to enhancing the reliability and precision of the proposed hybrid approach.

\input{image/experimentsetup}
\section{CONCLUSIONS}
This paper presented a novel DRL-based hybrid approach that combined deep reinforcement learning with adaptive scanning to automate radiotracer detection in robot-assisted radioguided surgeries. The adaptive scanning was used to provide an initial guess and pose for the DRL agent to facilitate robustness to variable target locations. The DRL was trained to utilise historical radio-guidance signals and proprioceptive states to optimise its direction to efficiently navigate toward the radio tracer. 
Simulation results validated the proposed method's superior accuracy, efficiency, and robustness compared to the individual methods applied separately. Real-world experiments further confirmed the feasibility of the hybrid method, successfully applying the algorithm to a real da Vinci robot system.

While the current experimental results show promising potential, future research will validate the approach through expanded trials and real radiotracer integration. By exploring diverse experimental conditions, this methodology advances automatic gamma probe radiotracer localization, bridging the gap between laboratory settings and realistic surgical procedures.



\section*{ACKNOWLEDGMENT}

This paper is independent research funded by the National Institute for Health Research (NIHR) Imperial Biomedical Research Centre (BRC), the Cancer Research UK (CRUK) Imperial Centre, the Wellcome Trust ITPA MedTechOne awards, and supported by Lightpoint Medical.

\newpage
\bibliographystyle{IEEEtran}
\bibliography{IEEEabrv,references}

\end{document}

%% file: image/problemstatement.tex
\begin{figure}
    \centering
    \includegraphics[width=1\linewidth]{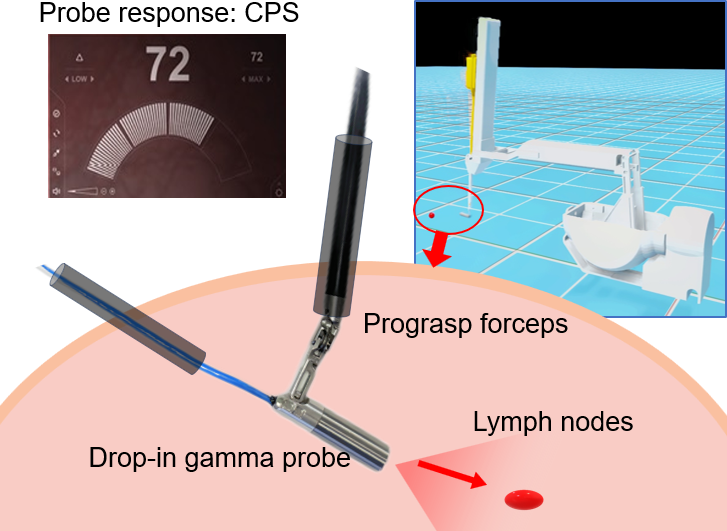}
    \caption{Process of radiotracer detection during robot-assisted surgery. The ProGrasp forceps hold the drop-in gamma probe, which is used to detect radioactive lymph nodes. As the probe moves closer to the target, the increase in real-time CPS feedback indicates the probe's proximity to the radiotracer. The precise manipulation of the probe thereby aids the localization of sentinel lymph nodes during procedures such as SLNB. }
    \label{fig:problemstatement}
\end{figure}

%% file: image/framework.tex
\begin{figure*}
    \centering
    \includegraphics[width=1\linewidth]{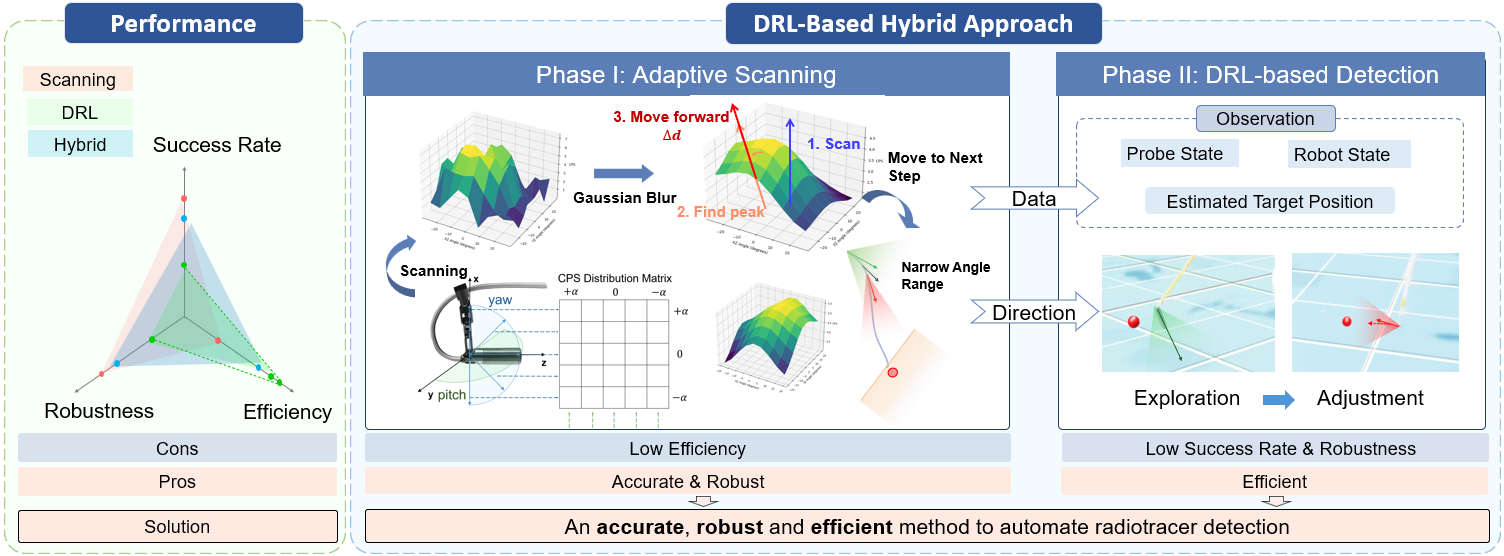}
    \caption{Overview of the proposed DRL-based hybrid approach for radiotracer detection. The method is divided into two phases: Phase I involves adaptive scanning to systematically detect the direction of target and narrow the detection range, while Phase II employs DRL for precise localization of the radiotracer. Phase I provides initial data for DRL to estimate a rough target position for state space at the beginning stage, as well as direction guidance to reduce exploration in DRL. The performance comparison shows that the hybrid method combines the accuracy and robustness of scanning with the efficiency of DRL, resulting in an accurate and efficient approach. The hybrid method achieves a balance between the reliability of classical scanning and the efficiency of DRL.}
    \label{fig:framework}
\end{figure*}

%% file: image/simulationenv.tex
\begin{figure}
    \centering
    \includegraphics[width=1\linewidth]{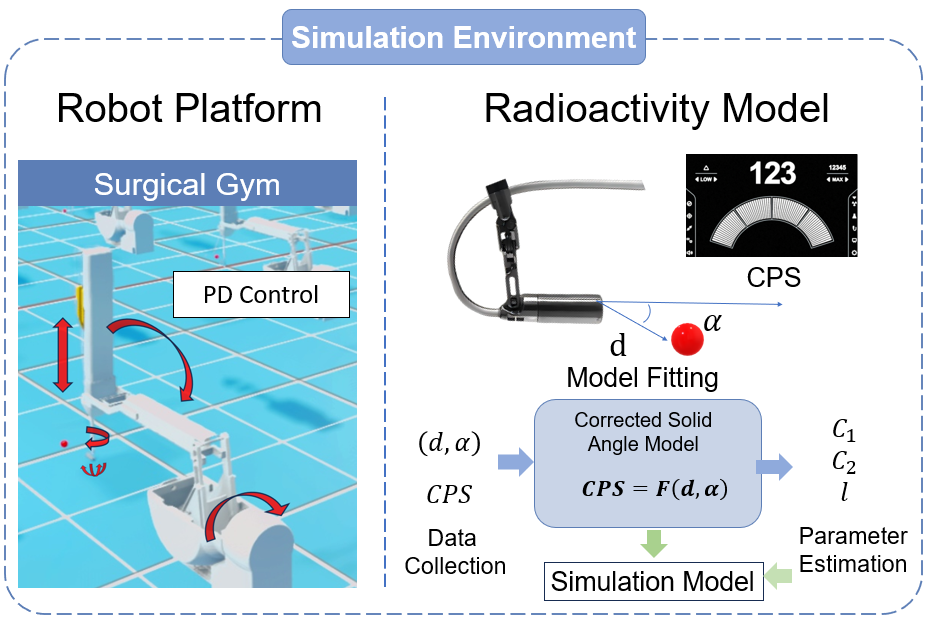}
    \caption{The simulation environment consists of a Robot Platform and a Radioactivity Model. The Robot Platform (left) utilizes Surgical Gym to simulate the da Vinci PSM with PD controllers for joint positions, enabling efficient training through GPU-accelerated simulations. The Radioactivity Model (right) uses the Corrected Solid Angle Model to simulate the gamma probe’s response and fitted through experimental data.}
    \label{fig:simulationenv}
\end{figure}

%% file: image/modelfitting_fig.tex
\begin{figure}
  \centering
    \includegraphics[width=0.9\linewidth]{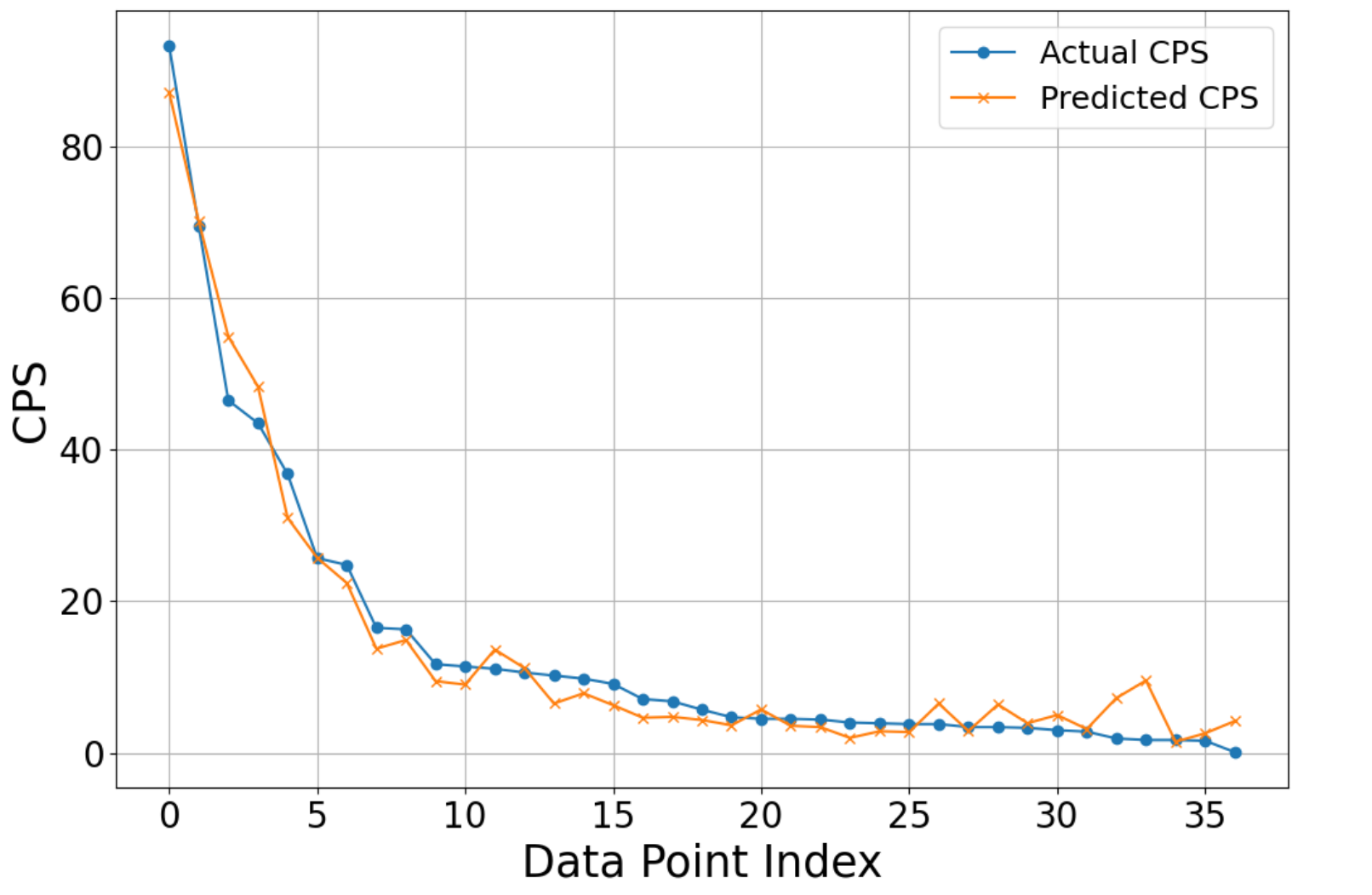}
    \caption{Comparison of Actual and Predicted CPS: the predicted CPS values from the model closely match the actual CPS values collected during the experiments, demonstrating the accuracy of the model. }
    \label{fig:modelfit}
\end{figure}

%% file: image/reward.tex
\begin{figure}
    \centering
    \includegraphics[width=1\linewidth]{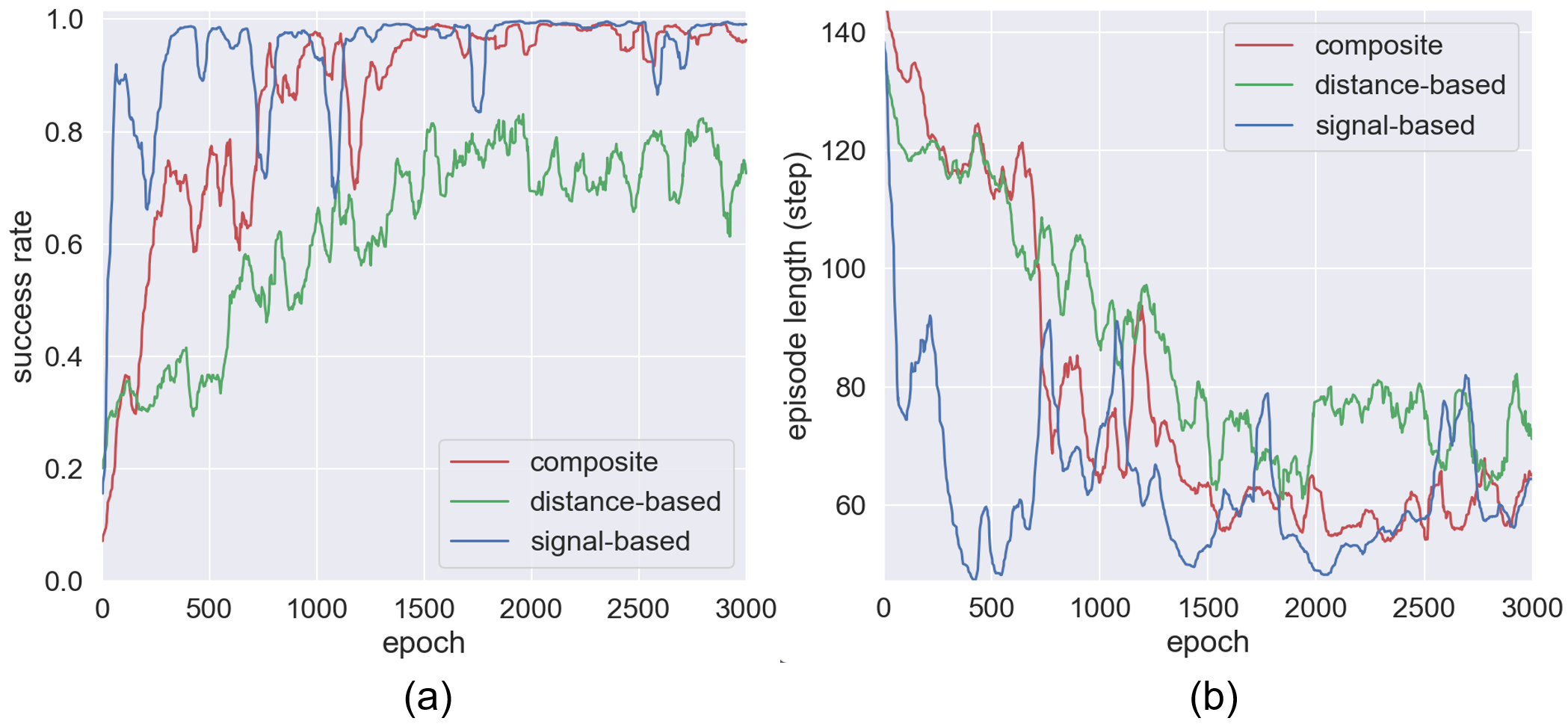}
    \caption{Comparison of three different reward settings on success rate (a) and episode length (b). The distance-based reward shows a lower success rate, longer episode, and slower learning, while the signal-based reward effectively guides the probe near the target but lacks stability. The composite approach achieves a balance between learning speed, stability, and overall performance.}
    \label{fig:success-rate}
\end{figure}

%% file: tables/robustness.tex
\begin{table}[hb]

\centering

\caption{Comparison of Robustness}
\scalebox{0.97}{

\begin{tabular}{ccccc}
\toprule
\multirow{2}{*}{\makecell[c]{\textbf{Randomization}\\ \textbf{range [mm]}}} & \multicolumn{3}{c}{\textbf{Success rate $\uparrow$}} \\ \cmidrule(lr){2-4}
 & \makecell[c]{\textbf{Adaptive}\\ \bf{scanning}} & \makecell[c]{\textbf{DRL}\\ \bf{approach}} & \makecell[c]{\textbf{Hybrid}\\ \bf{approach}} \\ \midrule
30    & 0.97                                   & 0.77           & \textbf{0.98} \\ 
50    & \textbf{0.96}                                   & 0.65           & 0.95 \\ 
70    & \textbf{0.94}                                   & 0.48           & 0.89 \\ \bottomrule
\end{tabular}

}
\label{tab:robust}
\end{table}

%% file: tables/performance.tex
\begin{table}[ht]
\centering

\caption{Comparison of Different Approaches}
\begin{tabular}{lcccc}
\toprule

\makecell[c]{\textbf{Evaluation}\\ \bf{metric}} & \makecell[c]{\textbf{Adaptive}\\ \bf {scanning}} & \makecell[c]{\textbf{DRL}\\ \bf{approach}} & \makecell[c]{\textbf{Hybrid}\\ \bf {approach}}\\ \hline
Average Steps $\downarrow$     & \textit{High}\textsuperscript{1}                  & \textbf{109}            & 95(+52)\textsuperscript{2} \\ 
Success Rate $\uparrow$     & \textbf{0.96}                                    & 0.65           & 0.95 \\ 
Robustness $\uparrow$\textsuperscript{3}        & \textbf{High}                   &Low              & \textbf{High} \\\bottomrule
\end{tabular}
\begin{flushleft}
\footnotesize
\textsuperscript{1} Marked as \textit{High} as it is significantly larger than other two approaches.\\

\textsuperscript{2} The 95 steps result from the DRL, while the additional 52 steps result from two rounds of scanning.\\
\textsuperscript{3} Robustness is evaluated based on the performance of each algorithm under varying levels of target position randomization (\autoref{tab:robust}).
\end{flushleft}
\label{tab:comparison}
\end{table}

%% file: image/experimentsetup.tex
\begin{figure}
    \centering

    \includegraphics[width=1\linewidth]{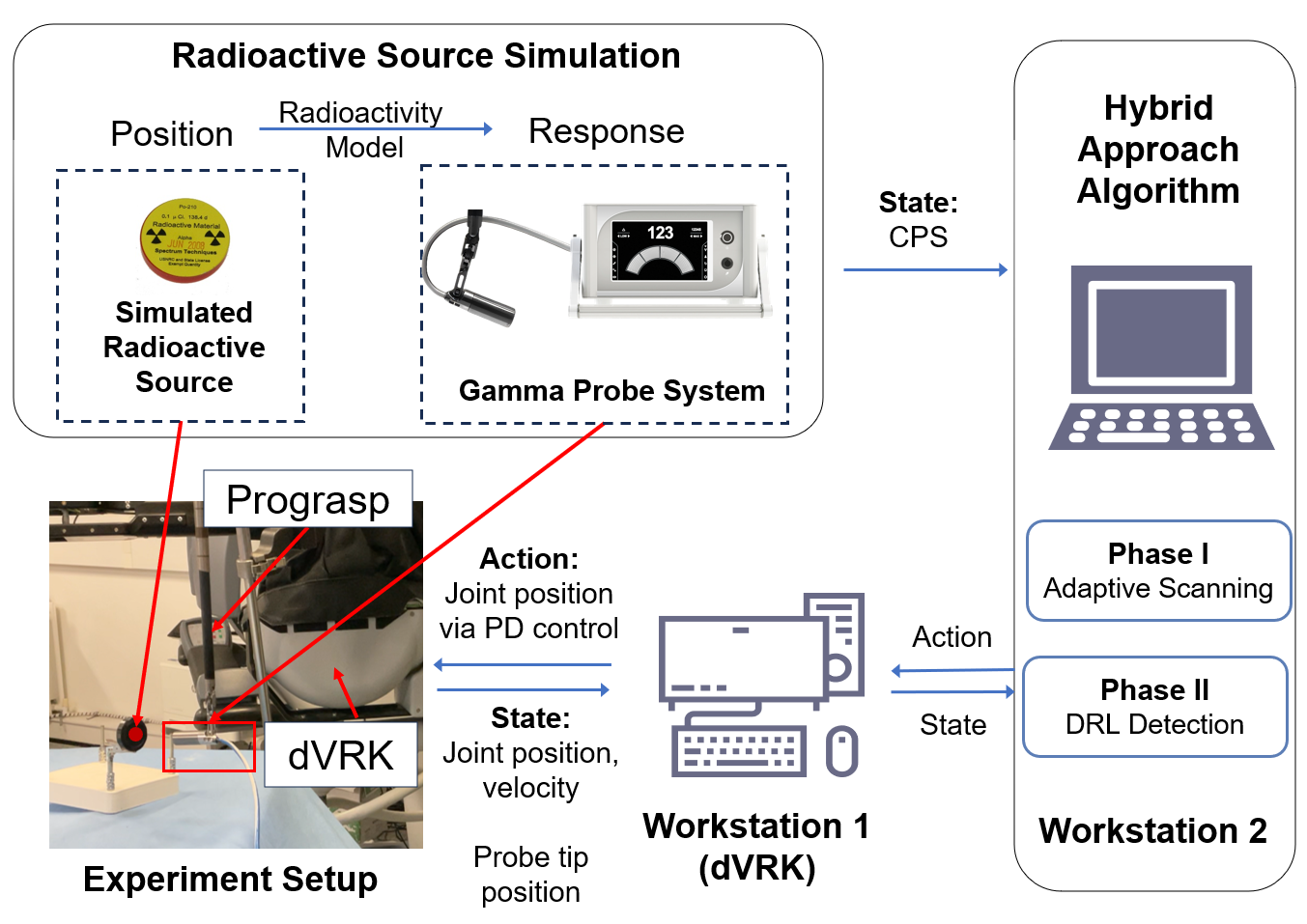}

    \caption{Illustration of the control and data flow between the da Vinci robot, the gamma probe system, and the two workstations. Workstation 1 communicates with the robot through the dVRK interface, while Workstation 2 runs the hybrid approach algorithm, receiving state data and sending control actions to guide the robot's movements toward the target.}
    \label{fig:experimentsetup}
\end{figure}